\documentclass[twoside]{article}

\usepackage[accepted]{aistats2017}
\usepackage[utf8]{inputenc} 
\usepackage[T1]{fontenc}    
\usepackage{url}            
\usepackage{booktabs}       
\usepackage{tabularx}
\usepackage{multirow}
\usepackage{amsfonts}       
\usepackage{nicefrac}       
\usepackage{microtype}      
\usepackage[round]{natbib}

\usepackage{lmodern}
\usepackage{amsmath}
\usepackage{mathtools}
\usepackage{amssymb}
\usepackage[hidelinks]{hyperref}
\usepackage{algorithm}
\usepackage[noend]{algpseudocode}
\usepackage{amsthm}
\usepackage{graphicx}
\usepackage{animate}
\usepackage{rotating}
\usepackage[colorinlistoftodos,textsize=tiny,textwidth=2cm]{todonotes}
\usepackage{dsfont}
\makeatletter
\newcounter{algorithmbis}
\def\BState{\State\hskip-\ALG@thistlm}
\renewcommand{\thealgorithmbis}{\arabic{algorithmbis}}
\def\algorithmbis{\@ifnextchar[{\@algorithmbisa}{\@algorithmbisb}}
\def\@algorithmbisa[#1]{%
  \refstepcounter{algorithmbis}
  \trivlist
  \leftmargin\z@
  \itemindent\z@
  \labelsep\z@
  \item[\parbox{\textwidth}{%
    \hrule
    \hrule
    \noindent\strut\textbf{Algorithm \thealgorithmbis} #1
    \hrule
  }]\hfil\vskip0em%
}
\def\@algorithmbisb{\@algorithmbisa[]}

\makeatother

\DeclareMathOperator{\E}{\mathbb{E}}

\newcommand{\KL}[2]{D_{\text{KL}}\left(#1 \mid\mid #2\right)}

\newcommand{\given}{\lvert}

\renewcommand{\vec}[1]{\boldsymbol{\mathbf{#1}}}

\usepackage{listings}
\usepackage{color}

\usepackage{bbm}

\usepackage{etoolbox}
\newcommand{\zerodisplayskips}{%
  \setlength{\abovedisplayskip}{6.5pt}%
  \setlength{\belowdisplayskip}{6.5pt}%
  \setlength{\abovedisplayshortskip}{6.5pt}%
  \setlength{\belowdisplayshortskip}{6.5pt}}
\appto{\normalsize}{\zerodisplayskips}
\appto{\small}{\zerodisplayskips}
\appto{\footnotesize}{\zerodisplayskips}

\begin{document}

\twocolumn[
    \aistatstitle{Inference Compilation and Universal Probabilistic Programming}
    \aistatsauthor{Tuan Anh Le \And Atılım Güneş Baydin \And Frank Wood}
    \aistatsaddress{
        Department of Engineering Science, University of Oxford \\
        \texttt{\{tuananh, gunes, fwood\}@robots.ox.ac.uk}
    }
]

\begin{abstract}
 We introduce a method for using  deep neural networks to amortize the cost of inference in models from the family induced by universal probabilistic programming languages, establishing a framework that combines the strengths of probabilistic programming and deep learning methods.  We call what we do ``compilation of inference'' because our method transforms a denotational specification of an inference problem in the form of a probabilistic program written in a universal programming language into a trained neural network denoted in a neural network specification language.  When at test time this neural network is fed observational data and executed, it performs approximate inference in the original model specified by the probabilistic program.  Our training objective and learning procedure are designed to allow the trained neural network to be used as a proposal distribution in a sequential importance sampling inference engine.  We illustrate our method on mixture models and Captcha solving and show significant speedups in the efficiency of inference.
\end{abstract}

\section{INTRODUCTION}

\begin{figure}[t]
    \centering
    \includegraphics[width=\linewidth]{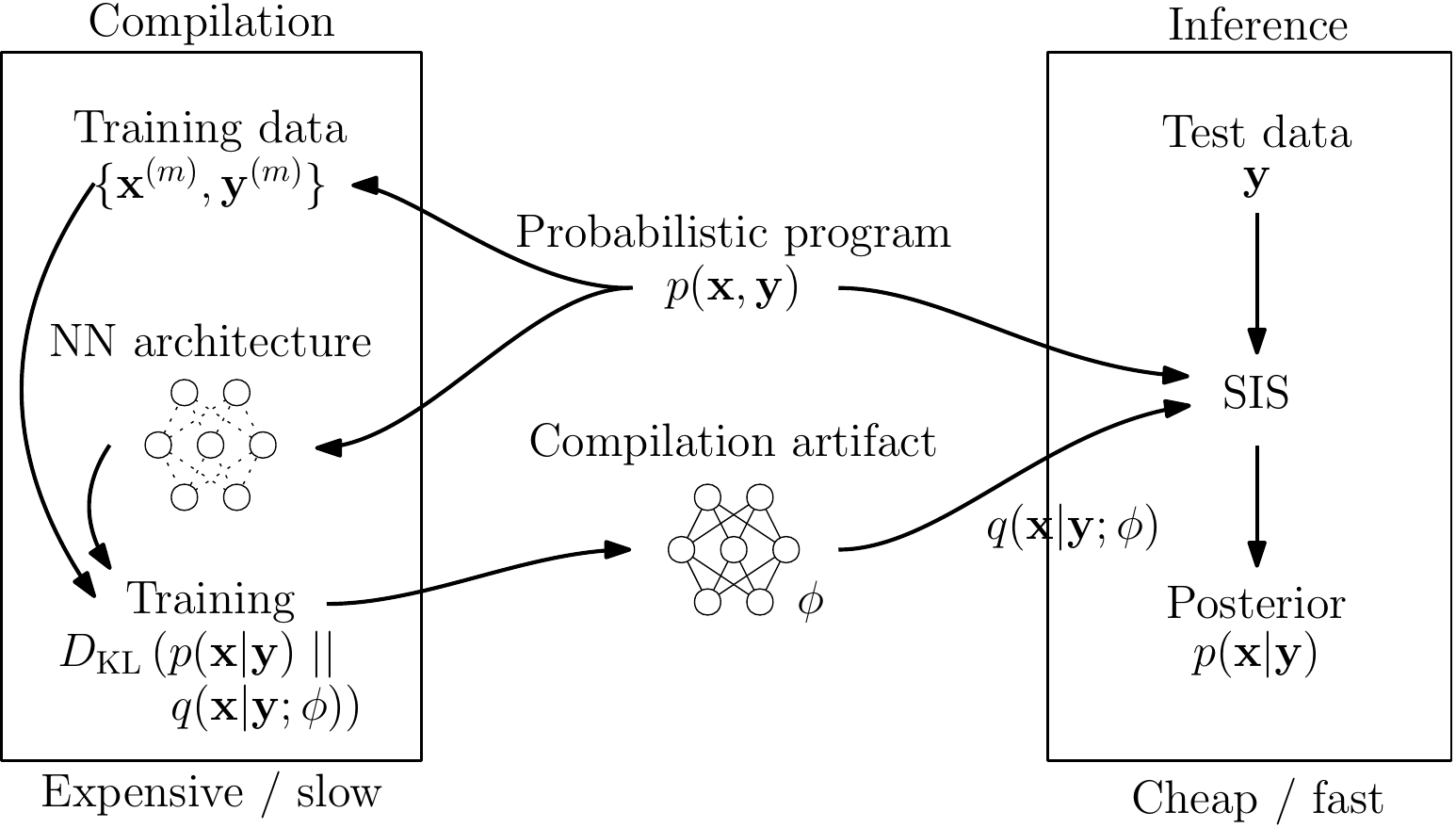}

    \caption{Our approach to compiled inference. Given only a probabilistic program $p(\mathbf{x}, \mathbf{y})$, during \emph{compilation} we automatically construct a neural network architecture comprising an LSTM core and various embedding and proposal layers specified by the probabilistic program and train this using an infinite stream of training data $\{\vec x^{(m)}, \vec y^{(m)}\}$ generated from the model. When this expensive compilation stage is complete, we are left with an artifact of weights $\phi$ and neural architecture specialized for the given probabilistic program. During \emph{inference}, the probabilistic program and the compilation artifact is used in a sequential importance sampling procedure, where the artifact parameterizes the proposal distribution $q(\mathbf{x} \given \mathbf{y}; \phi)$.}
    \label{fig:approach}
\end{figure}

Probabilistic programming uses computer programs to represent probabilistic models \citep{gordon2014probabilistic}. 
Probabilistic programming systems such as STAN \citep{carpenter2015stan}, BUGS \citep{lunn2000winbugs}, and Infer.NET \citep{InferNET14} allow efficient inference in a restricted space of generative models, while systems such as Church \citep{goodman2008church}, Venture \citep{mansinghka2014venture}, and Anglican \citep{wood2014new}---which we call \emph{universal}---allow inference in unrestricted models.
Universal probabilistic programming systems are built upon Turing complete programming languages which support constructs such as higher order functions, stochastic recursion, and control flow.

There has been a spate of recent work addressing the production of artifacts via ``compiling away'' or ``amortizing'' inference \citep{gershman2014amortized}.  This body of work is roughly organized into two camps. The one in which this work lives, arguably the camp organized around ``wake-sleep'' \citep{hinton1995wake},  is about offline unsupervised learning of observation-parameterized importance-sampling distributions for Monte Carlo inference algorithms.  In this camp,  the approach of \citet{paige2016inference} is closest to ours in spirit; they propose learning autoregressive neural density estimation networks offline that approximate inverse factorizations of graphical models so that at test time, the trained ``inference network''  starts with the values of all observed quantities and progressively proposes parameters for latent nodes in the original structured model.  However, inversion of the dependency structure is impossible in the universal probabilistic program model family,  so our approach instead focuses on learning proposals for ``forward'' inference methods in which no model dependency inversion is performed.  In this sense, our work can be seen as being inspired by that of \citet{kulkarni2015picture} and \citet{ritchie2016neurally} where program-specific neural proposal networks are trained to guide forward inference.  Our aim, though, is to be significantly less model-specific.  At a high level what characterizes this camp is the fact that the artifacts are trained to suggest sensible yet varied parameters for a given, explicitly structured and therefore potentially interpretable model.

The other related camp, emerging around the variational autoencoder \citep{kingma2013auto,burda2016importance}, also amortizes inference in the manner we describe, but additionally also simultaneously learns the generative model, within the structural regularization framework of a parameterized non-linear transformation of the latent variables. Approaches in this camp generally produce recognition networks that nonlinearly transform observational data at test time into parameters of a variational posterior approximation, albeit one with less conditional structure, excepting the recent work of \citet{johnson2016structured}.  A chief advantage of this approach is that the learned model, as opposed to the recognition network, is simultaneously regularized both towards being simple to perform inference in and towards explaining the data well.

In this work, we concern ourselves with performing inference in generative models specified as probabilistic programs while recognizing that alternative methods exist for amortizing inference while simultaneously learning model structure. Our contributions are twofold: (1) We work out ways to handle the complexities introduced when compiling inference for the class of generative models induced by universal probabilistic programming languages and establish a technique to embed neural networks in forward probabilistic programming inference methods such as sequential importance sampling \citep{doucet2009tutorial}. (2) We develop an adaptive neural network architecture, comprising a recurrent neural network core and embedding and proposal layers specified by the probabilistic program, that is reconfigured on-the-fly for each execution trace and trained with an infinite stream of training data sampled from the generative model. This establishes a framework combining deep neural networks and generative modeling with universal probabilistic programs (Figure~\ref{fig:approach}).

We begin by providing background information and reviewing related work in Section~\ref{sec:background}. In Section~\ref{sec:approach} we introduce inference compilation for sequential importance sampling, the objective function, and the neural network architecture. Section~\ref{sec:experiments} demonstrates our approach on two examples, mixture models and Captcha solving, followed by the discussion in Section~\ref{sec:discussion}.

\section{BACKGROUND}
\label{sec:background}

\subsection{Probabilistic Programming}
\label{sec:background/prob-prog}
Probabilistic programs denote probabilistic generative models as programs that include \texttt{sample} and \texttt{observe} statements \citep{gordon2014probabilistic}.
Both \texttt{sample} and \texttt{observe} are functions that specify random variables in this generative model using probability distribution objects as an argument, while \texttt{observe}, in addition, specifies the conditioning of this random variable upon a particular observed value in a second argument.
These observed values induce a conditional probability distribution over the execution traces whose approximations and expected values we want to characterize by performing inference.

An execution trace of a probabilistic program is obtained by successively executing the program deterministically, except when encountering \texttt{sample} statements at which point a value is generated according to the specified probability distribution and appended to the execution trace.
We assume the order in which the \texttt{observe} statements are encountered is fixed.
Hence we denote the observed values by $\vec y := (y_n)_{n = 1}^N$ for a fixed $N$ in all possible traces.

\begin{figure*}
    \centering
    \includegraphics[height=2.8cm]{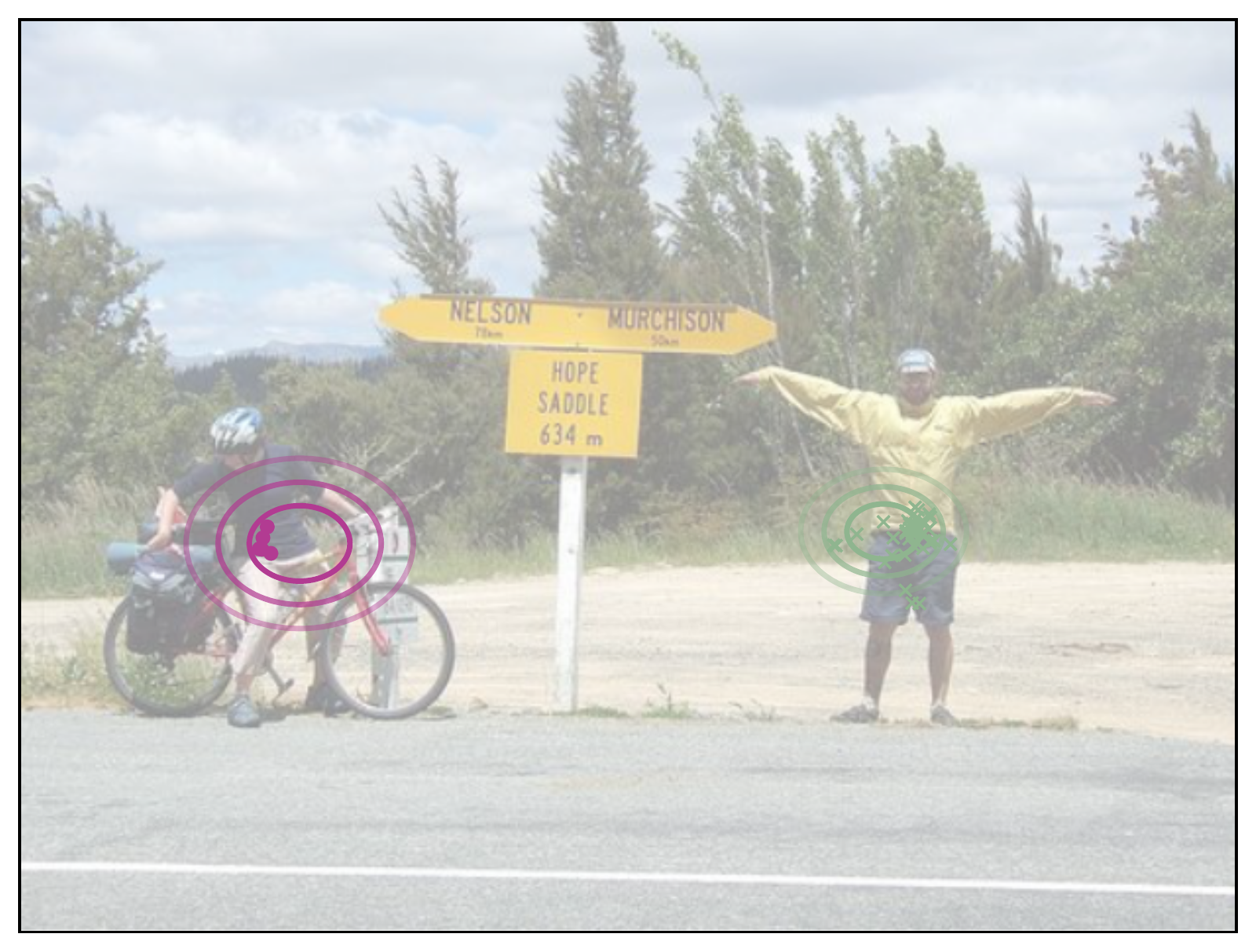}
    \includegraphics[height=2.8cm]{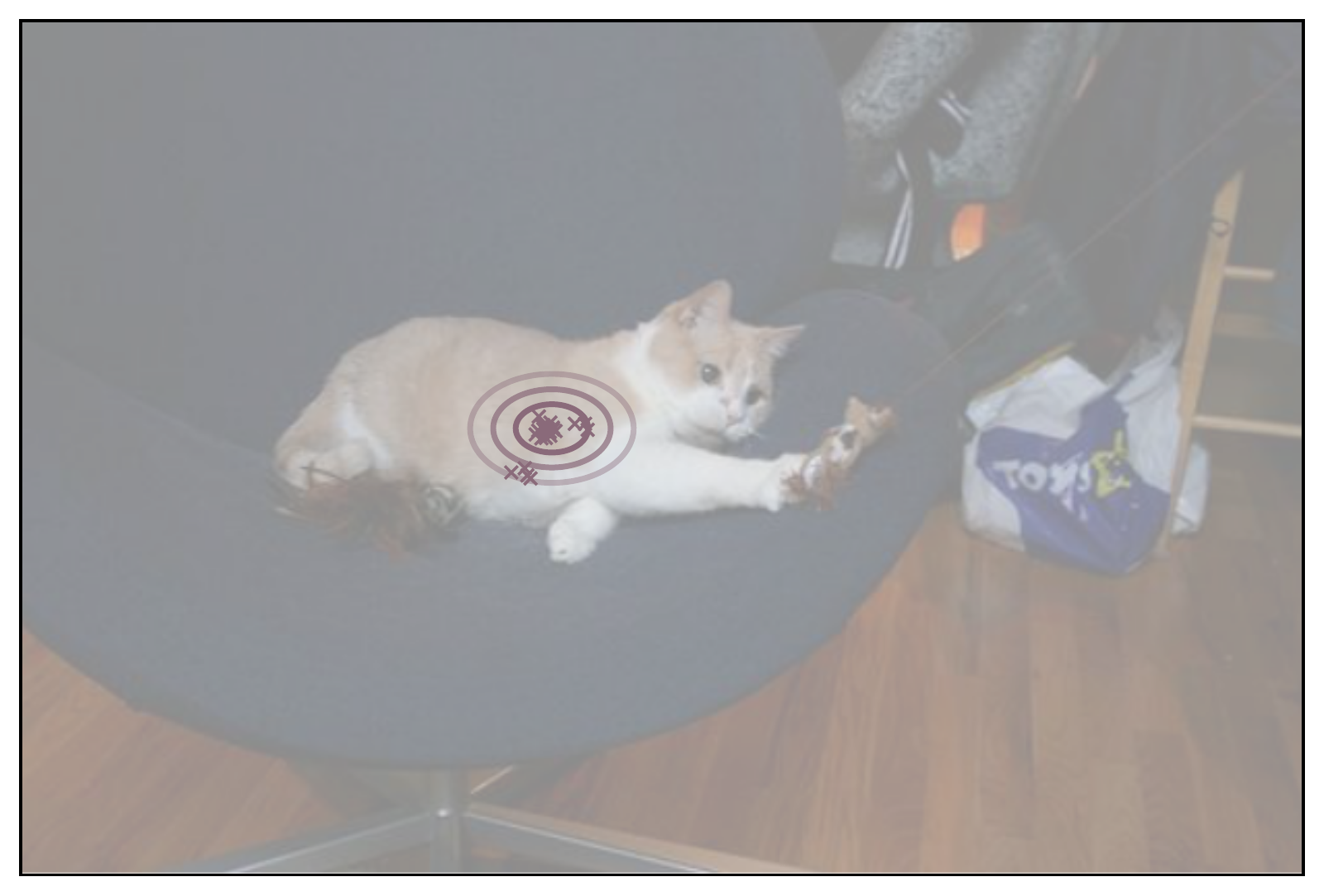}
    \includegraphics[height=2.8cm]{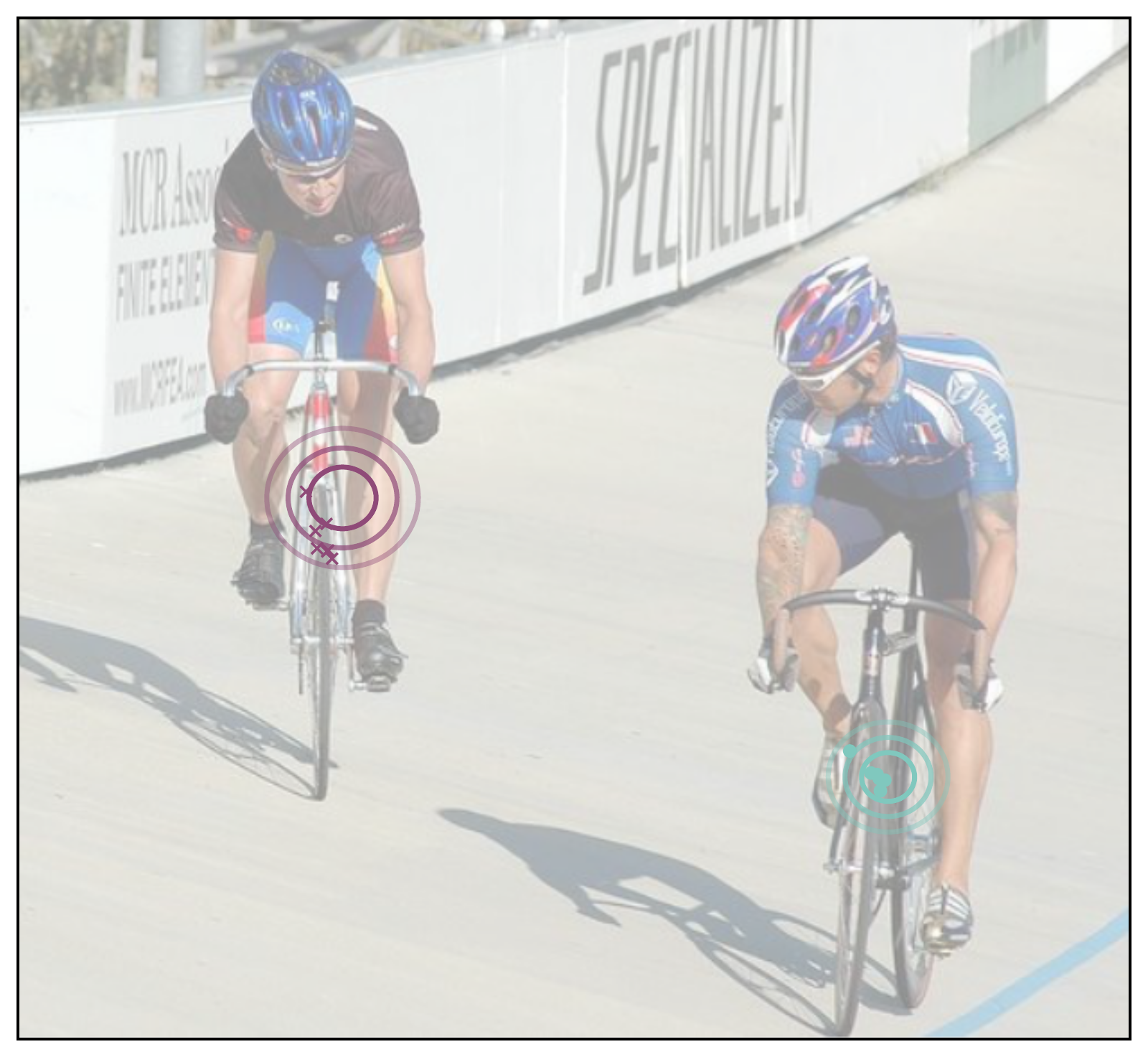}
    \includegraphics[height=2.8cm]{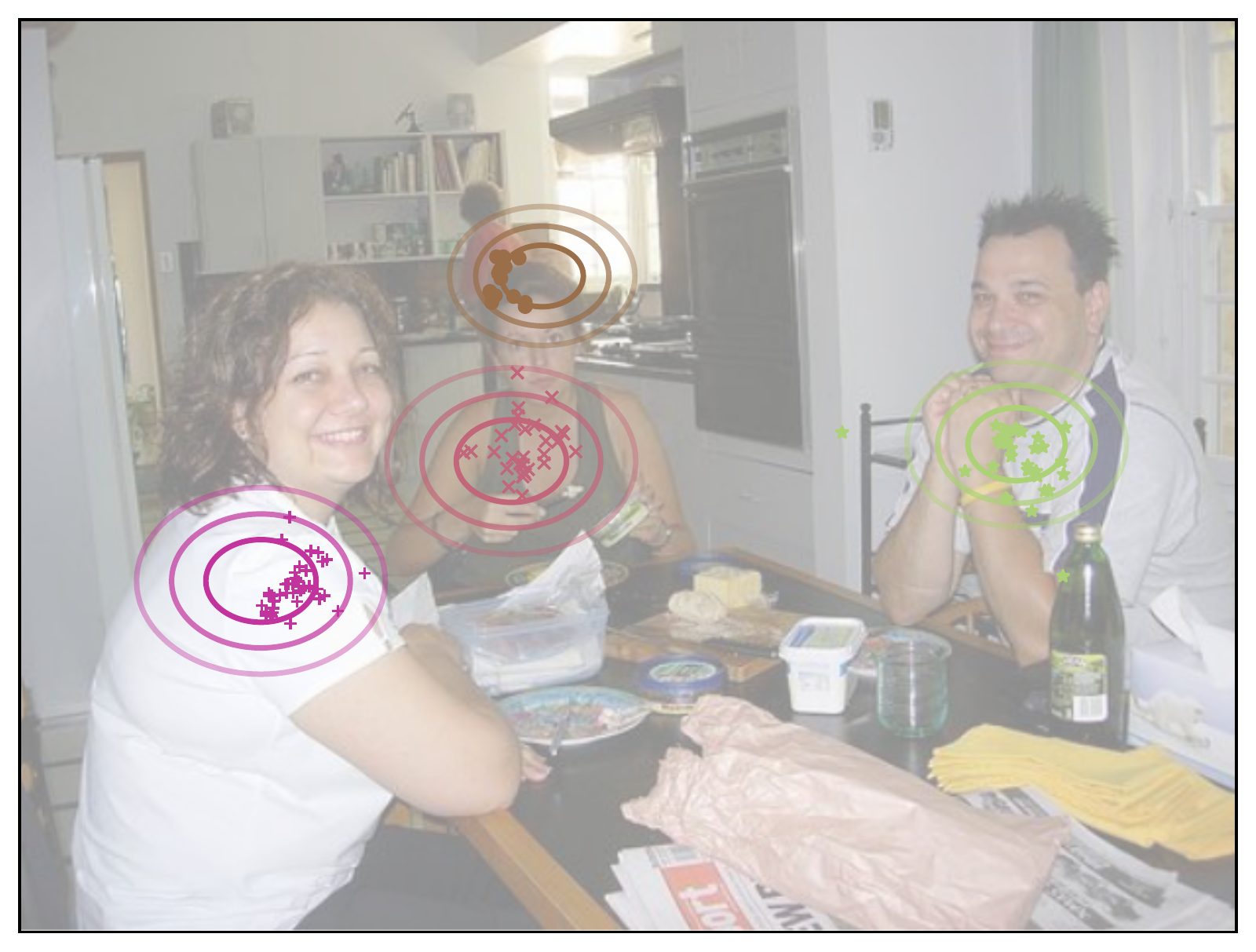}
    
    \caption{Results from counting and localizing objects detected in the PASCAL VOC 2007 dataset \citep{everingham2010pascal}.  We use the corresponding categories of object detectors (i.e., person, cat, bicycle) from the MatConvNet \citep{vedaldi15matconvnet} implementation of the Fast R-CNN \citep{girshick2015fast}.  The detector output is processed by using a high detection threshold and summarized by representing the bounding box detector output by a single central point.  Inference using a single trained neural network was able to accurately identify both the number of detected objects and their locations for all categories. MAP results from 100 particles.}
    \label{fig:object-counting-and-localization}
\end{figure*}
Depending on the probabilistic program and the values generated at \texttt{sample} statements, the order in which the execution encounters \texttt{sample} statements as well as the number of encountered \texttt{sample} statements may be different from one trace to another.
Therefore, given a scheme which assigns a unique address to each \texttt{sample} statement according to its lexical position in the probabilistic program, we represent an execution trace of a probabilistic program as a sequence 
\begin{align}
    \left(x_t, a_t, i_t\right)_{t=1}^{T}\;,
    \label{eq:background/trace}
\end{align}
where $x_t$, $a_t$, and $i_t$ are respectively the sample value, address, and instance (call number) of the $t$th entry in a given trace, and $T$ is a trace-dependent length. Instance values $i_t = \sum_{j=1}^t \mathds{1}(a_t = a_j)$ count the number of sample values obtained from the specific \texttt{sample} statement at address $a_t$, up to time step $t$.
For each trace, a sequence $\vec x := (x_t)_{t = 1}^T$ holds the $T$ sampled values from the \texttt{sample} statements.

The joint probability density of an execution trace is
\begin{align}
    p(\mathbf{x}, \mathbf{y}) := \prod_{t=1}^T f_{a_t}\left(x_t \given {x}_{1:t-1}\right) \prod_{n = 1}^N g_n(y_n \given {x}_{1:\tau(n)})\;, \label{eq:background/prob-prog/joint}
\end{align}
where $f_{a_t}$ is the probability distribution specified by the \texttt{sample} statement at address $a_t$ and $g_n$ is the probability distribution specified by the $n$th \texttt{observe} statement.
$f_{a_t}(\cdot \given x_{1:t - 1})$ is called the prior conditional density given the sample values $x_{1:t - 1}$ obtained before encountering the $t$th \texttt{sample} statement.
$g_n(\cdot \given x_{1:\tau(n)})$ is called the likelihood density given the sample values $x_{1:\tau(n)}$ obtained before encountering the $n$th \texttt{observe} statement, where $\tau$ is a mapping from the index $n$ of the \texttt{observe} statement to the index of the last \texttt{sample} statement encountered before this \texttt{observe} statement during the execution of the program.

Inference in such models amounts to computing an approximation of $p(\mathbf{x} \given \mathbf{y})$ and its expected values $I_{\zeta} = \int \zeta(\vec x) p(\vec x \given \vec y) \,\mathrm d\vec x$ over chosen functions $\zeta$. 

While there are many inference algorithms for universal probabilistic programming languages \citep{wingate2011lightweight,ritchie2015c3,wood2014new,paige2014asynchronous,rainforth2016interacting}, we focus on algorithms in the importance sampling family in the context of which we will develop our scheme for amortized inference.
This is related, but different to the approaches that adapt proposal distributions for the importance sampling family of algorithms \citep{gu2015neural,cheng2000ais}.

\subsection{Sequential Importance Sampling}
\label{sec:background/sis}

Sequential importance sampling (SIS) \citep{arulampalam2002tutorial,doucet2009tutorial} is a method for performing inference over execution traces of a probabilistic program \citep{wood2014new}
whereby a weighted set of samples $\{(w^k, \vec x^k)\}_{k = 1}^K$ is used to approximate the posterior and the expectations of functions as
\begin{align}
    \hat p(\vec x \given \vec y) &= \sum_{k = 1}^K w^k \delta(\vec x^k - \vec x) / \sum_{j = 1}^K w^j \\
    \hat I_{\zeta} &= \sum_{k = 1}^K w^k \zeta(\vec x^k) / \sum_{j = 1}^K w^j,
\end{align}
where $\delta$ is the Dirac delta function.

SIS requires designing proposal distributions $q_{a, i}$ corresponding to the addresses $a$ of all \texttt{sample} statements in the probabilistic program and their instance values $i$.
A proposal execution trace $x_{1:T^k}^k$ is built by executing the program as usual, except when a \texttt{sample} statement at address $a_t$ is encountered at time $t$, a proposal sample value $x_t^k$ is sampled from the proposal distribution $q_{a_t, i_t}(\cdot \given x_{1:t - 1}^k)$ given the proposal sample values until that point.
We obtain $K$ proposal execution traces $\vec x^k := x_{1:T^k}^k$ (possibly in parallel) to which we assign weights
\begin{align}
    w^k = \prod_{n = 1}^N g_n(y_n \given x_{1:\tau_k(n)}^k) \cdot \prod_{t = 1}^{T^k} \frac{f_{a_t}(x_t^k \given x_{1:t - 1}^k)}{q_{a_t, i_t}(x_t^k \given x_{1:t - 1}^k)}
\end{align}
for $k = 1, \dotsc, K$ with $T^k$ denoting the length of the $k$th proposal execution trace.

\section{APPROACH}
\label{sec:approach}

We achieve inference compilation in universal probabilistic programming systems through proposal distribution adaptation, approximating $p(\vec x \given \vec y)$ in the framework of SIS. Assuming we have a set of adapted proposals $q_{a_t, i_t}(x_t \given x_{1:t - 1}, \vec y)$ such that their joint $q(\vec x \given \vec y)$ is close to $p(\vec x \given \vec y)$, the resulting inference algorithm remains unchanged from the one described in Section~\ref{sec:background/sis}, except the replacement of $q_{a_t, i_t}(x_t \given x_{1:t - 1})$ by $q_{a_t, i_t}(x_t \given x_{1:t - 1}, \vec y)$.

Inference compilation amounts to minimizing a function, specifically the loss of a neural network architecture, which makes the proposal distributions good in the sense that we specify in Section~\ref{sec:approach/obj-fun}.
The process of generating training data for this neural network architecture from the generative model is described in Section~\ref{sec:approach/training}.
At the end of training, we obtain a compilation artifact comprising the neural network components---the recurrent neural network core and the embedding and proposal layers corresponding to the original model denoted by the probabilistic program---and the set of trained weights, as described in Section~\ref{sec:approach/nn-arch}.

\subsection{Objective Function}
\label{sec:approach/obj-fun}

We use the Kullback--Leibler divergence $\KL{p(\mathbf{x} \given \mathbf{y})}{q(\mathbf{x} \given \mathbf{y}; \phi)}$ as our measure of closeness between $p(\vec x \given \vec y)$ and $q(\vec x \given \vec y; \phi)$.
To achieve closeness over many possible $\vec y$'s, we take the expectation of this quantity under the distribution of $p(\mathbf{y})$ and ignore the terms excluding $\phi$ in the last equality:
\begin{align}
    \mathcal L(\phi) &:= \E_{p(\mathbf{y})}\left[\KL{p(\mathbf{x} \given \mathbf{y})}{q(\mathbf{x} \given \mathbf{y}; \phi)}\right]\\
    &= \int_{\mathbf{y}} p(\mathbf{y}) \int_{\mathbf{x}} p(\mathbf{x} \given \mathbf{y}) \log\frac{p(\mathbf{x} \given \mathbf{y})}{q(\mathbf{x} \given \mathbf{y}; \phi)} \,\mathrm d\mathbf{x} \,\mathrm d\mathbf{y} \nonumber\\
    &= \E_{p(\mathbf{x}, \mathbf{y})}\left[-\log q(\mathbf{x} \given \mathbf{y}; \phi)\right] + \text{const.} \label{eq:approach/obj-fun}
\end{align}
This objective function corresponds to the negative entropy criterion. Individual adapted proposals $q_{a_t, i_t}(x_t \given \eta_t(x_{1:t - 1}, \vec y, \phi)) =: q_{a_t, i_t}(x_t \given x_{1:t - 1}, \vec y)$ depend on $\eta_t$, the output of the neural network at time step $t$, parameterized by $\phi$.

Considering the factorization
\begin{align}
    q(\mathbf{x} \given \mathbf{y}; \phi) &= \prod_{t = 1}^T q_{a_t, i_t}(x_t \given \eta_t(x_{1:t - 1}, \vec y, \phi))\;,
\end{align}
the neural network architecture must be able to map to a variable number of outputs, and incorporate sampled values in a sequential manner, concurrent with the running of the inference engine.
We describe our neural network architecture in detail in Section~\ref{sec:approach/nn-arch}.

\subsection{Training Data}
\label{sec:approach/training}

Since Eq.~\ref{eq:approach/obj-fun} is an expectation over the joint distribution, we can use the following noisy unbiased estimate of its gradient to minimize the objective:
\begin{align}
    \frac{\partial}{\partial \phi}\mathcal L(\phi) \approx \frac{1}{M} \sum_{m = 1}^M \frac{\partial}{\partial \phi} \left( -\log q(\vec x^{(m)} \given \vec y^{(m)}; \phi) \right) \\
    (\vec x^{(m)}, \vec y^{(m)}) \sim p(\mathbf{x}, \mathbf{y}),\; m = 1, \dots, M\;.
\end{align}
Here, $(\vec x^{(m)}, \vec y^{(m)})$ is the $m$th training (probabilistic program execution) trace generated by running an unconstrained probabilistic program corresponding to the original one.
This unconstrained probabilistic program is obtained by a program transformation which replaces each \texttt{observe} statement in the original program by \texttt{sample} and ignores its second argument.

Universal probabilistic programming languages support stochastic branching and can generate execution traces with a changing (and possibly unbounded) number of random choices.
We must, therefore, keep track of information about the addresses and instances of the samples $x_t^{(m)}$ in the execution trace, as introduced in Eq.~\ref{eq:background/trace}.
Specifically, we generate our training data in the form of minibatches \citep{cotter2011better} sampled from the generative model $p(\vec x, \vec y)$:
\begin{align}
    & \mathcal{D}_{\textrm{train}} = \left\{\left(x_t^{(m)}, a_t^{(m)}, i_t^{(m)}\right)_{t=1}^{T^{(m)}}, \left(y_n^{(m)}\right)_{n=1}^{N} \right\}_{m=1}^{M}\;,
\end{align}
where $M$ is the minibatch size, and, for a given trace $m$, the sample values, addresses, and instances are respectively denoted $x_t^{(m)}$, $a_t^{(m)}$, and $i_t^{(m)}$, and the values sampled from the distributions in \texttt{observe} statements are denoted $y_n^{(m)}$.

During compilation, training minibatches are generated on-the-fly from the probabilistic generative model and streamed to a stochastic gradient descent (SGD) procedure, specifically Adam \citep{kingma2014adam}, for optimizing the neural network weights $\phi$.

Minibatches of this infinite stream of training data are discarded after each SGD update; we therefore have no notion of a finite training set and associated issues such as overfitting to a set of training data and early stopping using a validation set \citep{prechelt1998early}. We do sample a validation set that remains fixed during training to compute validation losses for tracking the progress of training in a less noisy way than that admitted by the training loss.

\subsection{Neural Network Architecture}
\label{sec:approach/nn-arch}

Our compilation artifact is a collection of neural network components and their trained weights, specialized in performing inference in the model specified by a given probabilistic program. The neural network architecture comprises a non-domain-specific recurrent neural network (RNN) core and domain-specific observation embedding and proposal layers specified by the given program. We denote the set of the combined parameters of all neural network components $\phi$.

RNNs are a popular class of neural network architecture which are well-suited for sequence-to-sequence modeling \citep{sutskever2014sequence} with a wide spectrum of state-of-the-art results in domains including machine translation \citep{bahdanau2014neural}, video captioning \citep{venugopalan2014translating}, and learning execution traces \citep{reed2015neural}. We use RNNs in this work owing to their ability to encode dependencies over time in the hidden state.
In particular, we use the long short-term memory (LSTM) architecture which helps mitigate the vanishing and exploding gradient problems of RNNs \citep{hochreiter1997long}.

\begin{figure}
    \centering
    \includegraphics[width=0.95\columnwidth]{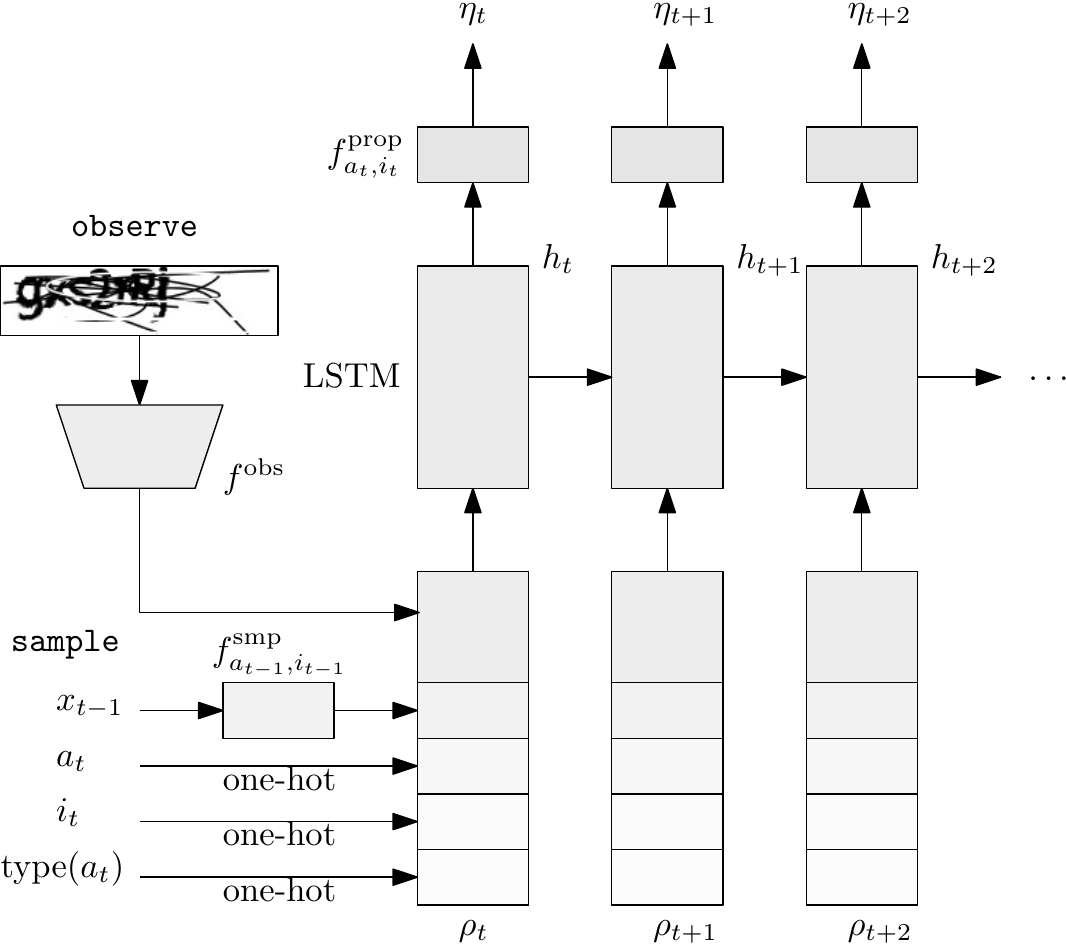}
    \caption{The neural network architecture. $f^\textrm{obs}$: \texttt{observe} embedding; $f_{a_{t-1},i_{t-1}}^{\textrm{smp}}$: sample embeddings; $x_{t-1}$: previous \texttt{sample} value; $a_t$, $i_t$, $\textrm{type}(a_t)$: one-hot encodings of current address, instance, proposal type; $\rho_t$: LSTM input; $h_t$: LSTM output; $f_{a_t, i_t}^{\textrm{prop}}$: proposal layers; $\eta_t$: proposal parameters. Note that the LSTM core can possibly be a stack of multiple LSTMs.}
    \label{fig:nn-arch}
\end{figure}

The overall architecture (Figure~\ref{fig:nn-arch}) is formed by combining the LSTM core with a domain-specific \texttt{observe} embedding layer $f^{\textrm{obs}}$, and several \texttt{sample} embedding layers $f_{a, i}^{\textrm{smp}}$ and proposal layers $f_{a, i}^{\textrm{prop}}$ that are distinct for each address--instance pair $(a, i)$. As described in Section~\ref{sec:approach/training}, each probabilistic program execution trace can be of different length and composed of a different sequence of addresses and instances. To handle this complexity, we define an adaptive neural network architecture that is reconfigured for each encountered trace by attaching the corresponding embedding and proposal layers to the LSTM core, creating new layers on-the-fly on the first encounter with each $(a, i)$ pair.

Evaluation starts by computing the \texttt{observe} embedding $f^{\textrm{obs}}\left(\vec y\right)$. This embedding is computed once per trace and repeatedly supplied as an input to the LSTM at each time step. Another alternative is to supply this embedding only once in the first time step, an approach preferred by \citet{karpathy2015deep} and \citet{vinyals2015show} to prevent overfitting (also see Section~\ref{sec:experiments/captcha}). 

At each time step $t$, the input $\rho_t$ of the LSTM is constructed as a concatenation of 
\begin{enumerate}
    \item the \texttt{observe} embedding $f^{\textrm{obs}}(\vec y)$,
    \item the embedding of the previous \texttt{sample} $f_{a_{t-1}, i_{t-1}}^{\textrm{smp}}\left(x_{t-1}\right)$, using zero for $t=1$, and
    \item the one-hot encodings of the current address $a_t$, instance $i_t$, and proposal type $\textrm{type}(a_t)$ of the \texttt{sample} statement
\end{enumerate}
for which the artifact will generate the parameter $\eta_t$ of the proposal distribution $q_{a_t, i_t}(\cdot \given \eta_t)$. The parameter $\eta_t$ is obtained via the proposal layer $f_{a_t, i_t}^{\textrm{prop}}\left(h_t\right)$, mapping the LSTM output $h_t$ through the corresponding proposal layer. The LSTM network has the capacity to incorporate inputs in its hidden state.
This allows the parametric proposal $q_{a_t, i_t}(x_t \given \eta_t(x_{1:t - 1}, \vec y, \phi))$ to take into account all previous samples and all observations.

During training (compilation), we supply the actual sample values $x_{t - 1}^{(m)}$ to the embedding $f_{a_{t-1}, i_{t-1}}^{\textrm{smp}}$, and we are interested in the parameter $\eta_t$ in order to calculate the per-sample gradient $\frac{\partial}{\partial \phi} - \log q_{a_t^{(m)}, i_t^{(m)}}(x_t^{(m)} \given \eta_t(x_{1:t - 1}, \vec y, \phi))$ to use in SGD. 

During inference, the evaluation proceeds by requesting proposal parameters $\eta_t$ from the artifact for specific address--instance pairs $(a_t, i_t)$ as these are encountered. The value $x_{t-1}$ is sampled from the proposal distribution in the previous time step.

The neural network artifact is implemented in Torch \citep{collobert2011torch7}, and it uses a ZeroMQ-based protocol for interfacing with the Anglican probabilistic programming system \citep{wood2014new}.  This setup allows distributed training (e.g., \citet{dean2012largescale}) and inference with GPU support across many machines, which is beyond the scope of this paper. The source code for our framework and for reproducing the experiments in this paper can be found on our project page.\footnote{\url{https://probprog.github.io/inference-compilation/}}

\section{EXPERIMENTS}
\label{sec:experiments}

We demonstrate our inference compilation framework on two examples.
In our first example we demonstrate an open-universe mixture model. 
In our second, we demonstrate Captcha solving via probabilistic inference \citep{mansinghka2013approximate}.\footnote{A video of inference on real test data for both examples is available at: \url{https://youtu.be/m-FYEXVyQjQ}}

\begin{figure*}
    \centering
    \includegraphics[width=0.95\textwidth,trim={3mm 3mm 4mm 2mm}]{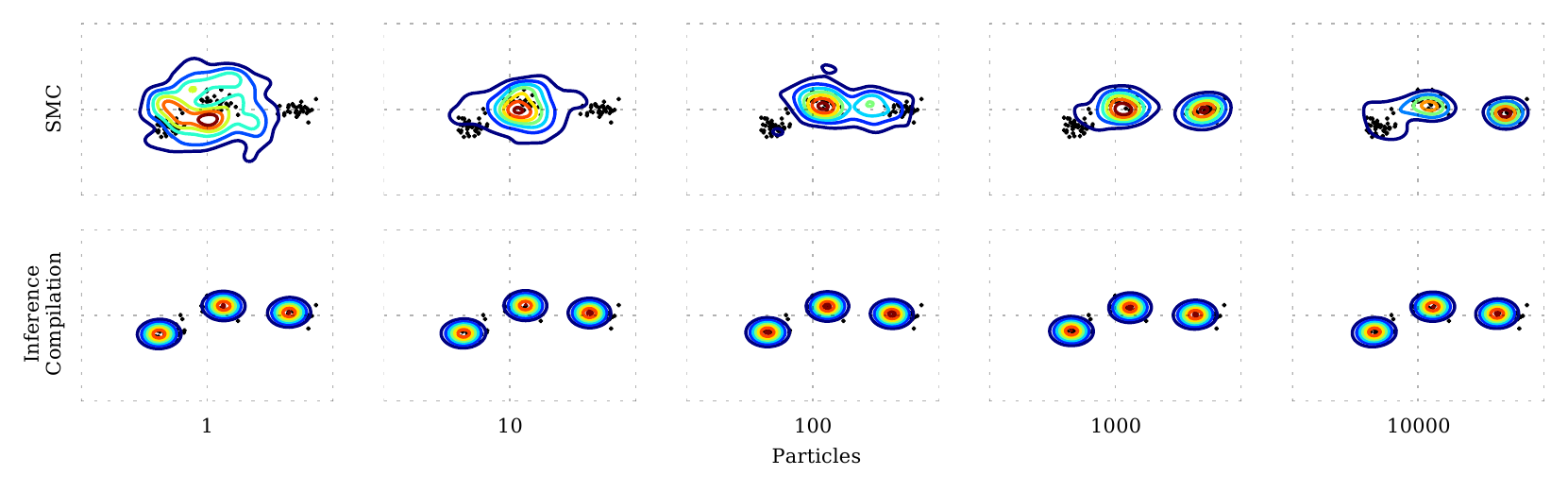}
    \caption{Typical inference results for an isotropic Gaussian mixture model with number of clusters fixed to $K=3$.
    Shown in all panels: kernel density estimation of the distribution over maximum a posteriori values of the means $\{\max_{\mu_k} p(\mu_k \given \vec y)\}_{k = 1}^3$ over 50 independent runs.
    This figure illustrates the uncertainty in the estimate of where cluster means are  for each given number of particles, or equivalently, fixed amount of computation.
    The top row shows that, given more computation, inference, as expected, slowly becomes less noisy in expectation.
    In contrast, the bottom row shows that the proposal learned and used by inference compilation produces a low-noise, highly accurate estimate given even a very small amount of computation.
    Effectively, the encoder learns to simultaneously localize all of the clusters highly accurately.}
    \label{fig:gmm-2d}
\end{figure*}

\subsection{Mixture Models}
\label{sec:experiments/gmm}

Mixture modeling, e.g.~the Gaussian mixture model (GMM) shown in Figure~\ref{fig:gmm}, is about density estimation, clustering, and counting.  The inference problems posed by a GMM, given a set of vector observations, are to identify how many, where, and how big the clusters are, and optionally, which data points belong to each cluster.  

We investigate inference compilation for a two-dimensional GMM in which the number of clusters is unknown.  Inference arises from observing the values of $y_n$ (Figure~\ref{fig:gmm}, line 9) and inferring the posterior number of clusters $K$ and the set of cluster mean and covariance parameters $\{\mu_k,\Sigma_k\}_{k=1}^K$.  We assume that the input data to this model has been translated to the origin and normalized to lie within $[-1,1]$ in both dimensions.

In order to make good proposals for such inference, the neural network must be able to count, i.e., extract and represent information about how many clusters there are and, conditioned on that, to localize the clusters.  Towards that end, we select a convolutional neural network as the observation embedding, whose input is a two-dimensional histogram image of binned observed data $\vec y$.

In presenting observational data $\vec y$ assumed to arise from a mixture model to the neural network, there are some important considerations that must be accounted for.  In particular, there are symmetries in mixture models \citep{nishihara2013detecting} that must be broken in order for training and inference to work.  First, there are $K!$ (factorial) ways to label the classes.  Second, there are $N!$ ways the individual data points could be permuted.  Even in experiments like ours with $K<6$ and $N\approx100$, this presents a major challenge for neural network training.  We break the first symmetry by, at training time, sorting the clusters by the Euclidian distance of their means from the origin and relabeling all points with a permutation that labels points from the cluster nearest the original as coming from the first cluster, next closest the second, and so on.  This is only approximately symmetry breaking as many different clusters may be very nearly the same distance away from the origin.  Second, we avoid the $N!$ symmetry by only predicting the number, means, and covariances of the clusters, not the individual cluster assignments. 
The net effect of the sorting is that the proposal mechanism will learn to propose the nearest cluster to the origin as it receives training data always sorted in this manner.  

Figure~\ref{fig:gmm-2d}, where we fix the number of clusters to $3$, shows that we are able to learn a proposal that makes inference dramatically more efficient than sequential Monte Carlo (SMC) \citep{doucet2009tutorial}.  Figure~\ref{fig:object-counting-and-localization} shows one kind of application such an efficient inference engine can do: simultaneous object counting \citep{lempitsky2010learning} and localization for computer vision, where we achieve counting by setting the prior $p(K)$ over number of clusters to be a uniform distribution over $\{1, 2, \dotsc, 5\}$.

\setlength{\fboxsep}{0pt}
\begin{figure}
    \centering
    \scriptsize

    \begin{algorithmic}[1]
    \Procedure{GaussianMixture}{}
    \State $K \sim p(K)$\Comment \texttt{sample} number of clusters
    \For{$k = 1, \dots, K$}
    \State $\vec \mu_k, \vec \Sigma_k \sim p(\vec \mu_k,\vec \Sigma_k)$\Comment \texttt{sample} cluster parameters
    \EndFor

    \BState \emph{Generate data}:
    \State $\vec \pi \gets $uniform$(1,K)$
    \For{$n = 1, \dots, N$}
    \State $z_n \sim p(z_n | \vec \pi)$\Comment \texttt{sample} class label
    \State $y_n \sim p(y_n | z_n=k, \vec \mu_k, \vec \Sigma_k)$\Comment \texttt{sample} data
    \EndFor
    
    \State {\textbf return} $y_n$
    \EndProcedure

    \end{algorithmic}
    \vspace{2mm}
    \caption{Pseudo algorithm for generating Gaussian mixtures of a variable number of clusters. At test time we \texttt{observe} data $y_n$ and infer $K,\{\vec \mu_k, \vec \Sigma_k\}_{k=1}^K$.}
    \label{fig:gmm}
\end{figure}

\subsection{Captcha Solving}
\label{sec:experiments/captcha}

We also demonstrate our inference compilation framework by writing generative probabilistic models for Captchas \citep{von2003captcha} and comparing our results with the literature. Captcha solving is well suited for a generative probabilistic programming approach because its latent parameterization is low-dimensional and interpretable by design. Using conventional computer vision techniques, the problem has been previously approached using segment-and-classify pipelines \citep{starostenko2015breaking,bursztein2014end,gao2014robustness,gao2013robustness}, and state-of-the-art results have been obtained by using deep convolutional neural networks (CNNs) \citep{GoodfellowBIAS13,stark-gcpr15}, at the cost of requiring very large (in the order of millions) labeled training sets for supervised learning.

\setlength{\fboxsep}{0pt}
\begin{figure}[t]
    \centering
    \scriptsize

    \begin{algorithmic}[1]
    \Procedure{Captcha}{}
    \State $\nu \sim p(\nu)$
    \Comment \texttt{sample} number of letters
    \State $\kappa \sim p(\kappa)$
    \Comment \texttt{sample} kerning value

    \BState \emph{Generate letters}:
    \State $\Lambda \gets \{\}$
    \For{$i = 1, \dots, \nu$}
    \State $\lambda \sim p(\lambda)$
    \Comment \texttt{sample} letter identity
    \State $\Lambda \gets \textrm{append}(\Lambda, \lambda)$
    \EndFor

    \BState \emph{Render}:
    \State $\gamma \gets \textrm{render}(\Lambda, \kappa)$
    \State $\pi \sim p(\pi)$
    \Comment \texttt{sample} noise parameters
    \State $\gamma \gets \textrm{noise}(\gamma, \pi)$
    
    \State {\textbf return} $\gamma$
    \EndProcedure
    \end{algorithmic}
    \vspace{2mm}

    \begin{tabular}{@{}p{0.25\columnwidth}@{}p{0.25\columnwidth}@{}p{0.25\columnwidth}@{}p{0.25\columnwidth}@{}}
    \fbox{\includegraphics[width=0.23\columnwidth]{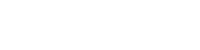}} &
    \fbox{\includegraphics[width=0.23\columnwidth]{facebook-0.png}} &
    \fbox{\includegraphics[width=0.23\columnwidth]{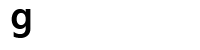}} &
    \fbox{\includegraphics[width=0.23\columnwidth]{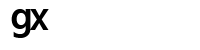}} \\
    $a_1=\,$``$\nu$'' & $a_2=\,$``$\kappa$'' & $a_3=\,$``$\lambda$'' & $a_4=\,$``$\lambda$''\\
    $i_1=1$ & $i_2=1$ & $i_3=1$ & $i_4=2$\\
    $x_1 = 7$ & $x_2 = -1$ & $x_3 = 6$ & $x_4 = 23$\\[1mm]
    \fbox{\includegraphics[width=0.23\columnwidth]{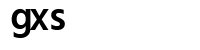}} &
    \fbox{\includegraphics[width=0.23\columnwidth]{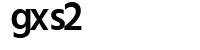}} &
    \fbox{\includegraphics[width=0.23\columnwidth]{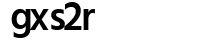}} &
    \fbox{\includegraphics[width=0.23\columnwidth]{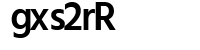}} \\
    $a_5=\,$``$\lambda$'' & $a_6=\,$``$\lambda$'' & $a_7=\,$``$\lambda$'' & $a_8=\,$``$\lambda$''\\
    $i_5=3$ & $i_6=4$ & $i_7=5$ & $i_8=6$\\
    $x_5 = 18$ & $x_6 = 53$ & $x_7 = 17$ & $x_8 = 43$\\[1mm]
    \fbox{\includegraphics[width=0.23\columnwidth]{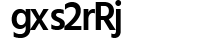}} &
    \fbox{\includegraphics[width=0.23\columnwidth]{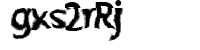}} &
    \fbox{\includegraphics[width=0.23\columnwidth]{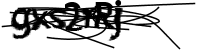}} &
    \fbox{\includegraphics[width=0.23\columnwidth]{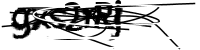}}\\
    $a_9=\,$``$\lambda$'' & Noise: & Noise: & Noise: \\
    $i_9 = 7$ & displacement & stroke & ellipse\\
    $x_9 = 9$ & field\\
    \end{tabular}
    \caption{
        Pseudo algorithm and a sample trace of the Facebook Captcha generative process.
        Variations include sampling font styles, coordinates for letter placement, and language-model-like letter identity distributions $p(\lambda \given \lambda_{1:t-1})$ (e.g., for meaningful Captchas).
        Noise parameters $\pi$ may or may not be a part of inference.
        At test time we \texttt{observe} image $\gamma$ and infer $\nu,\Lambda$.}
    \label{fig:captcha}
\end{figure}

We start by writing generative models for each of the types surveyed by \citet{bursztein2014end}, namely Baidu 2011 (\includegraphics[height=3mm]{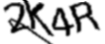}), Baidu 2013 (\includegraphics[height=3mm]{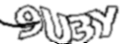}), eBay (\includegraphics[height=3mm]{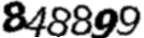}), Yahoo (\includegraphics[height=3mm]{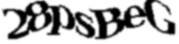}), reCaptcha (\includegraphics[height=3mm]{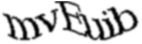}), and Wikipedia (\includegraphics[height=3mm]{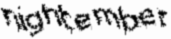}). Figure~\ref{fig:captcha} provides an overall summary of our modeling approach. The actual models include domain-specific letter dictionaries, font styles, and various types of renderer noise for matching each Captcha style. In particular, implementing the displacement fields technique of \citet{simard2003} proved instrumental in achieving our results. Note that the parameters of stochastic renderer noise are not inferred in the example of Figure~\ref{fig:captcha}. Our experiments have shown that we can successfully train artifacts that also extract renderer noise parameters, but excluding these from the list of addresses for which we learn proposal distributions improves robustness when testing with data not sampled from the same model. This corresponds to the well-known technique of adding synthetic variations to training data for transformation invariance, as used by \citet{simard2003}, \citet{varga2003generation}, \citet{jaderberg2014synthetic}, and many others. 

For the compilation artifacts we use a stack of two LSTMs of 512 hidden units each, an \texttt{observe}-embedding CNN consisting of six convolutions and two linear layers organized as [2$\times$Convolution]-MaxPooling-[3$\times$Convolution]-MaxPooling-Convolution-MaxPooling-Linear-Linear, where convolutions are 3$\times$3 with successively 64, 64, 64, 128, 128, 128 filters, max-pooling layers are 2$\times$2 with step size 2, and the resulting embedding vector is of length 1024. All convolutions and linear layers are followed by ReLU activation. Depending on the particular style, each artifact has approximately 20M trainable parameters.
Artifacts are trained end-to-end using Adam \citep{kingma2014adam} with initial learning rate $\alpha = 0.0001$, hyperparameters $\beta_1 = 0.9$, $\beta_2 = 0.999$, and minibatches of size 128. 

\begin{table*}
  \footnotesize
  \setlength{\tabcolsep}{1mm}
  \caption{Captcha recognition rates.}
  \label{table:captcha-results}
  \def\arraystretch{1.15}
  \begin{tabularx}{\textwidth}{@{}lXXXXXXX@{}}
    \toprule
    & Baidu 2011 & Baidu 2013 & eBay & Yahoo & reCaptcha & Wikipedia & Facebook \\
    \midrule
    Our method & 99.8\% & 99.9\% & 99.2\% & 98.4\% & 96.4\% & 93.6\% & 91.0\%\\
    \citeauthor{bursztein2014end}~(\citeyear{bursztein2014end}) & 38.68\% & 55.22\% & 51.39\% & 5.33\% & 22.67\% & 28.29\% \\
    \citeauthor{starostenko2015breaking}~(\citeyear{starostenko2015breaking}) & & & & 91.5\% & 54.6\% & \\
    \citeauthor{gao2014robustness}~(\citeyear{gao2014robustness}) & 34\% & & & 55\% & 34\% & & \\
    \citeauthor{gao2013robustness}~(\citeyear{gao2013robustness}) & & 51\% & & 36\% \\
    \citeauthor{GoodfellowBIAS13}~(\citeyear{GoodfellowBIAS13}) & & & & & 99.8\%\\
    \citeauthor{stark-gcpr15}~(\citeyear{stark-gcpr15}) & & & & & 90\% \\
    \bottomrule
\end{tabularx}
\end{table*}

Table~\ref{table:captcha-results} reports inference results with test images sampled from the model, where we achieve very high recognition rates across the board. The reported results are obtained after approximately 16M training traces. With the resulting artifacts, running inference on a test Captcha takes $<$~100~ms, whereas durations ranging from 500~ms \citep{starostenko2015breaking} to 7.95~s \citep{bursztein2014end} have been reported with segment-and-classify approaches.
We also compared our approach with the one by \citet{mansinghka2013approximate}.
Their method is slow since it must be run anew for each Captcha, taking in the order of minutes to solve one Captcha in our implementation of their method.
The probabilistic program must also be written in a way amenable to Markov Chain Monte Carlo inference such as having auxiliary indicator random variables for rendering letters to overcome multimodality in the posterior.

We subsequently investigated how the trained models would perform on Captcha images collected from the web. We identified Wikipedia and Facebook as two major services still making use of textual Captchas, and collected and labeled test sets of 500 images each.\footnote{Facebook Captchas are collected from a page for accessing groups. Wikipedia Captchas appear on the account creation page.} Initially obtaining low recognition rates ($<$ 10\%), with several iterations of model modifications (involving tuning of the prior distributions for font size and renderer noise), we were able to achieve 81\% and 42\% recognition rates with real Wikipedia and Facebook datasets, considerably higher than the threshold of 1\% needed to deem a Captcha scheme broken \citep{bursztein2011text}. The fact that we had to tune our priors highlights the issues of model bias and ``synthetic gap'' \citep{zhang2015learning} when training models with synthetic data and testing with real data.\footnote{Note that the synthetic/real boundary is not always clear: for instance, we assume that the Captcha results in \citet{GoodfellowBIAS13} closely correspond to our results with synthetic test data because the authors have access to Google's true generative process of reCaptcha images for their synthetic training data. \citet{stark-gcpr15} both train and test their model with synthetic data.}

In our experiments we also investigated feeding the \texttt{observe} embeddings to the LSTM at all time steps versus only in the first time step. We empirically verified that both methods produce equivalent results, but the latter takes significantly (approx. 3 times) longer to train. This is because we are training $f^{\textbf{obs}}$ end-to-end from scratch, and the former setup results in more frequent gradient updates for $f^{\textbf{obs}}$ per training trace.\footnote{Both \citet{karpathy2015deep} and \citet{vinyals2015show}, who feed CNN output to an RNN only once, use pretrained embedding layers.}

In summary, we only need to write a probabilistic generative model that produces Captchas sufficiently similar to those that we would like to solve. Using our inference compilation framework, we get the inference neural network architecture, training data, and labels for free. If you can create instances of a Captcha, you can break it.

\section{DISCUSSION}
\label{sec:discussion}

We have explored making use of deep neural networks for amortizing the cost of inference in probabilistic programming. In particular, we transform an inference problem given in the form of a probabilistic program into a trained neural network architecture that parameterizes proposal distributions during sequential importance sampling. The amortized inference technique presented here provides a framework within which to integrate the expressiveness of universal probabilistic programming languages for generative modeling and the processing speed of deep neural networks for inference. This merger addresses several fundamental challenges associated with its constituents: fast and scalable inference on probabilistic programs, interpretability of the generative model, an infinite stream of labeled training data, and the ability to correctly represent and handle uncertainty. 

Our experimental results show that, for the family of models on which we focused, the proposed neural network architecture can be successfully trained to approximate the parameters of the posterior distribution in the \texttt{sample} space with nonlinear regression from the \texttt{observe} space. There are two aspects of this architecture that we are currently working on refining. Firstly, the structure of the neural network is not wholly determined by the given probabilistic program: the invariant LSTM core maintains long-term dependencies and acts as the glue between the embedding and proposal layers that are automatically configured for the address--instance pairs $(a_t, i_t)$ in the program traces. We would like to explore architectures where there is a tight correspondence between the neural artifact and the computational graph of the probabilistic program. Secondly, domain-specific \texttt{observe} embeddings such as the convolutional neural network that we designed for the Captcha-solving task are hand picked from a range of fully-connected, convolutional, and recurrent architectures and trained end-to-end together with the rest of the architecture. Future work will explore automating the selection of potentially pretrained embeddings.

A limitation that comes with not learning the generative model itself---as is done by the models organized around the variational autoencoder \citep{kingma2013auto,burda2016importance}---is the possibility of model misspecification \citep{shalizi2009dynamics,gelman2013philosophy}.
Section~\ref{sec:approach/training} explains that our training setup is exempt from the common problem of overfitting to the training set. But as demonstrated by the fact that we needed alterations in our Captcha model priors for handling real data, we do have a risk of overfitting to the model. Therefore we need to ensure that our generative model is ideally as close as possible to the true data generation process and remember that misspecification in terms of broadness is preferable to a misspecification where we have a narrow, but uncalibrated, model.

\subsubsection*{Acknowledgements}
We would like to thank Hakan Bilen for his help with the MatConvNet setup and showing us how to use his Fast R-CNN implementation and Tom Rainforth for his helpful advice. Tuan Anh Le is supported by EPSRC DTA and Google (project code DF6700) studentships. Atılım Güneş Baydin and Frank Wood are supported under DARPA PPAML through the U.S. AFRL under Cooperative Agreement FA8750-14-2-0006, Sub Award number 61160290-111668.

\renewcommand\refname{\vskip -1cm}
\subsubsection*{References}
\bibliographystyle{abbrvnat}
\bibliography{comp-inf}

\end{document}